\documentclass[conference]{IEEEtran}
\IEEEoverridecommandlockouts
\pdfoutput=1
\usepackage{cite}
\usepackage{amsmath,amssymb,amsfonts}
\usepackage{algorithmic}
\usepackage{graphicx}
\usepackage{textcomp}
\usepackage{xcolor}
\usepackage{tabularx}
\def\BibTeX{{\rm B\kern-.05em{\sc i\kern-.025em b}\kern-.08em
    T\kern-.1667em\lower.7ex\hbox{E}\kern-.125emX}}
\usepackage{dblfloatfix} % floats in twocolumns come out in the right order
\begin{document}

\title{The impact of Twitter on political influence on the choice of a running mate: Social Network Analysis and Semantic Analysis - A Review\\
}

\author{\IEEEauthorblockN{Immaculate Wanza Musyoka}
\IEEEauthorblockA{\textit{dept. Computer Science} \\
\textit{Dedan Kimathi University of Technology} \\
immaculate.musyoka@dkut.ac.ke}
\and
\IEEEauthorblockN{Irad Kamuti Mwendo}
\IEEEauthorblockA{\textit{dept. Computer Science} \\
\textit{Dedan Kimathi University of Technology}\\
iradspm@gmail.com }
\and
\IEEEauthorblockN{David Gichohi Maina}
\IEEEauthorblockA{\textit{dept. Computer Science} \\
\textit{Dedan Kimathi University of Technology }\\
david.gichohi@dkut.ac.ke}
\and
\IEEEauthorblockN{Kinyua Gikunda}
\IEEEauthorblockA{\textit{dept. Computer Science} \\
\textit{Dedan Kimathi University of Technology}\\
patrick.gikunda@dkut.ac.ke}
}

\maketitle

\begin{abstract}
In this new era of social media, social networks are becoming increasingly important sources of user-generated content on the internet. These kinds of information resources, which include a lot of people's feelings, opinions, feedback, and reviews, are very useful for big businesses, markets, politics, journalism, and many other fields. Politics is one of the most talked-about and popular topics on social media networks right now. Many politicians use micro-blogging services like Twitter because they have a large number of followers and supporters on those networks. Politicians, political parties, political organizations, and foundations use social media networks to communicate with citizens ahead of time. Today, social media is used by hundreds of thousands of political groups and politicians. On these social media networks, every politician and political party has millions of followers, and politicians find new and innovative ways to urge individuals to participate in politics. Furthermore, social media assists politicians in various decision-making processes by providing recommendations, such as developing policies and strategies based on previous experiences, recommending and selecting suitable candidates for a particular constituency, recommending a suitable person for a particular position in the party, and launching a political campaign based on citizen sentiments on various issues and controversies, among other things.
This research is a review on the use of social network analysis (SNA) and semantic analysis (SA) on the Twitter platform to study the supporters’ networks of political leaders because it can help in decision-making when predicting their political futures.

\end{abstract}

\begin{IEEEkeywords}
politician, running mate, supporter, social network analysis, semantic analysis.
\end{IEEEkeywords}

\section{Introduction}
Social media platforms are now a part of everyday life \cite{b1}. They let people from all over the world talk to each other in large groups about anything, even important social and political issues. So, social media has become a good source of data-rich information that can be used to learn about people's political opinions and interests \cite{b2}. Twitter, a medium for political and social discussion, can be analyzed to get political insights from people \cite{b1} .These insights are useful in making informed decisions. They help predict how a certain running mate will affect the chances of a political aspirant \cite{b3}.

Social network and semantic analysis are vital tools when it comes to the use of social media to make predictions \cite{b4}. Structures of interactions between individuals (or other social entities, such as organizations) and inter dependencies in behavior or attitudes connected to configurations of social relations are studied using social network analysis. A social network is a social structure made up of a group of social actors (such as people or organizations), a set of dyadic links, and other social interactions between them. It brings individuals together to converse, share ideas and interests, and meet new people. Social media is the term for this form of collaboration and sharing \cite{b5}. The social network approach offers a collection of tools for examining the structure of entire social entities, as well as a number of hypotheses to explain the patterns that emerge. Social network analysis is used to detect local and global trends, locate influential entities, and investigate network dynamics in the study of these systems.

Semantic networks are a type of data representation that incorporates linguistic information to express the relationship or dependency between concepts or objects. Semantic networks aid in the extraction of meanings given to a political leader by followers based on the text they use in social media posts and could help political leaders transform their strategies \cite{b6}. Semantic networks, rather than connecting individuals to people, connect words to words based on their co-occurrence \cite{b7}.Words that come in the same sentence as a politician's name may highlight areas of concern that require public relations assistance \cite{b5}.

Semantic networks differ from social network analysis in that nodes are semantic concepts (e.g., names, places, organizations, policies, values) rather than social actors (e.g., citizens, nongovernmental organizations [NGOs], political parties); and ties are not social interactions (e.g., friendships) but associations between concepts (e.g., co-occurrence). Semantic networks, on the other hand, can be studied using network analysis methodologies to gain quantitative and qualitative insights \cite{b5}. Quantitative measurements can aid qualitative semantic structure research and exploration \cite{b7}. This research is therefore guided by the following objectives:
1.	What different techniques have been used in SNA and SA in relation to politics?
2.	Which are the best-performing techniques in the analysis of SNA and SA?
3.	How can we apply the best techniques in SNA and SA to determine the political influence on the choice of a running mate?
\begin{figure}[htbp]
\centerline{\includegraphics[width=0.5\textwidth]{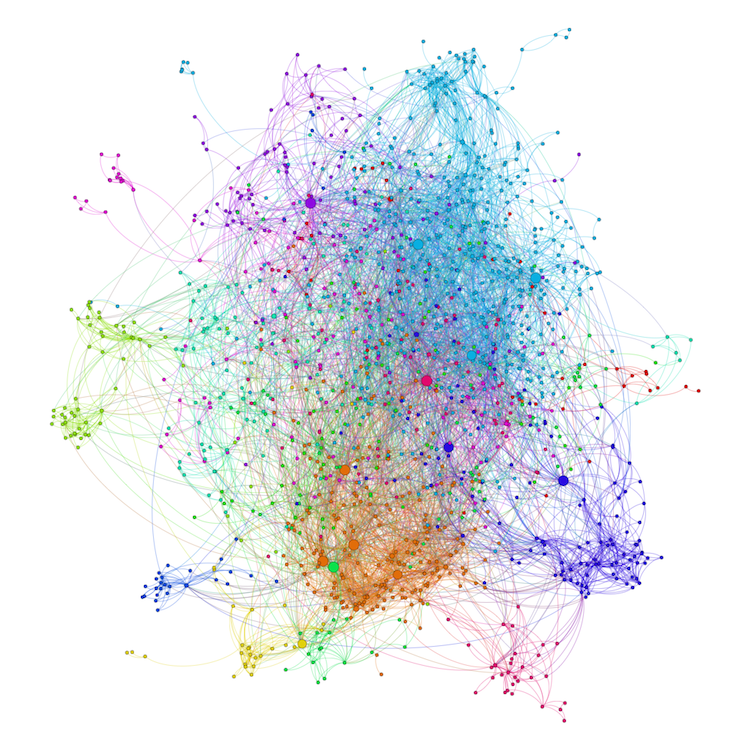}}
\caption{Example SNA network of tweets[8]}
\label{fig}
\end{figure}\\

The rest of the work is arranged as follows: in Section 2, an overview of the current state-of-the-art in social network analysis and semantic analysis is presented; and in Section 3, we provide the methodology for the best technique that can be used in social network analysis and sentiment analysis before making our conclusion in Section 4.

\section{State-of-the-art approaches for social network analysis and sentiment analysis}

In this section, we review various approaches that has been used in social network analysis and sentiment analysis. Most researchers have analyzed social media in different application areas \cite{b9}. For instance, \cite{b10} analyzed Twitter to predict 2020 Oscar winner using several machine learning techniques, \cite{b11} created a supervised learning algorithm to examine the political sentiments of India's national political parties with respect to the 2019 elections and \cite{b12} mined opinions on the web among others as discussed below.

A study done by \cite{b13} employed Naïve Bayes classifier to analyze opinions and categorize positive and negative comments on Facebook regarding agritech startups of Thailand to examine the sentiments and attitudes of people and investors. The researcher eliminated slang or unstructured words to reduce errors in the results and achieved an accuracy of 79. In the prediction of sentiments on comments made on Twitter, \cite{b14} manually tagged twitter comments as positive, neutral, and negative. Using Naïve Bayes to classify unknown comments, they achieved an accuracy score of 86.43 which was higher compared with decision tree and random forest algorithms. However, \cite{b15} classified Facebook textual posts of bigger size into negative, positive, neutral, spam, and dual classes ensuring that every post can be easily classified into its respective class. In doing so, an accuracy score of 91.2 was obtained. Research conducted using multinomial Naïve bayes to detect cyberstalking on social media, achieved an accuracy of 95.8 \cite{b16}. The researcher represented all tweets with bag-of-words treating each word as an independent feature before tagging. This variant of Naïve bayes proved to produce good results with proper feature extraction, hence the high accuracy. There is long-term semantic popularity of a post as compared to the social users’ popularity (likes count of a post). Using Gaussian Naïve Bayes, \cite{b17} achieved an average accuracy score of 90. This implies that appropriate future extraction and more label classes can be used to accurately predict new instances of sentiments.

Graph theories can be used in the analysis of content posted on social networks \cite{b18}. How users are connected to the content posted on social media can be easily modeled. To achieve higher predictions on the links between users and content posted on social networks, graph-related metrics such as betweenness, centrality, and clustering algorithms \cite{b19} can be used to easily group social networks into groups based on their links and relationships. For instance, \cite{b20} used the K-means clustering algorithm to analyze the sentiment polarity on threatening trends in Colombia. Furthermore, \cite{b21} use other algorithms like Fruchterman Reingold \cite{b22}, Harel-Korean, and Caluset-Newman-Moore to measure betweenness, closeness centralities to measure how politicians are engaged with citizens in developing countries using Twitter and in the automatic prediction of political views, \cite{b23} used constructed graph algorithms to forecast Russian elections in 2018. 

Supervised machine learning algorithms can be used to model political social network popularity. \cite{b3} in trying to predict the strongest and most influential supporters network of three famous Pakistan political leaders, they used Botometer to detect and remove fake supporters from the network before analyzing the retweet network and the reply network using SVM which achieved an accuracy of 79.89. \cite{b24} tried to predict the spread of propaganda in social media, they collected lots of textual data from different online sources, annotated it, and performed feature engineering of the data. Applying the output after feature engineering to Support Vector Machine (SVM) algorithm, an accuracy of 81 was attained. Comparing this result with \cite{b25} work who attained 88 accuracy after widening their network by combining the output of sentiment analysis of tweets and retweet count network and applying these results to the SVM algorithm. 

In detecting and identifying hate speech on social media comments, \cite{b26} considered social media sites such as Twitter, Facebook and Myspace. Collected corpus was pre-processed through stemming and normalization before applying the cleaned dataset on Random Forest classifier which achieved an accuracy of 72.22. In detecting depression and applying the same algorithm, \cite{b27} achieved an accuracy of 74 without feature selection despite having more features like tweets, hashtags, retweets and user information which were more than \cite{b26}. To analyze sentiments on Indonesia language, \cite{b28} used Random forest classifier and explored bag of words using short term weighting methods for words’ selection after which an accuracy of 82.9 was attained. \cite{b29} applied supervised ensemble random forest approach to detect cyber-bullying of children. In this case, the researcher used tweets and Facebook posts which were later labelled and preprocessed before being applied to the said approach, which attained 83 accuracy and 88 when carried out with larger dataset by \cite{b30}. Comparing these results with \cite{b31}, who integrated unigram and random forest to identify critical indicators of the rise and decline of stock market attained accuracy of 98.34.

There were two types of methods that can be used to analyze social networks:  content analysis methods and methods based on the analysis of graph structure topologies. Machine learning are used for content analysis while graph algorithms for topology analysis \cite{b18}. Though there are several machine learning techniques, SVM and Nave Bayes models are most frequently utilized. When applied to well-formed text corpora, naive Bayes is effective \cite{b32}, whereas support vector machines perform well on datasets with irregular shapes. However, machine learning methods struggle on social media platforms where users post in random lengths, slang or unstructured words, and with lots of spelling mistakes, and it takes a sizable training sample to modify the method because the size and caliber of the output are affected by the dataset size \cite{b33}. Additionally, using machine learning for analysis requires a lot of time, sometimes hours in complex models, especially if training is necessary \cite{b34}. A smaller training dataset speeds up the process, but the classification accuracy suffers \cite{b35}.

\begin{table*}[ht]
  \centering
  \resizebox{\textwidth}{!}{\begin{tabular}{*{4}{c}}
    \hline
    &\textbf{Problem} &\textbf{Approach} &\textbf{Performance and Reference}\\
    \hline
   1 &To examine the sentiments and attitudes of people and investors using Facebook Textual comments data & Naive Bayes & 79\% \cite{b13} \\
    \hline
   2 &Prediction of sentiments on comments made on Twitter  &Naive Bayes &86.43\% \cite{b14}  \\
    \hline
   3 &To predict the strongest and most influential supporters’ in Twitter  &NB, K-NN, and SVM &SVM 79.89\% NB 75.40\% and KNN 65.35\% \cite{b3}\\
    \hline
   4 &Classification Facebook textual posts \& Textual data-Facebook  &Naive Bayes &Accuracy 91.2\% \cite{b15} \\
    \hline  
   5 &To detect cyberstalking on social media \& Textual data- Twitter  &Multinomial Naive Bayes &Accuracy 95.8\% \cite{b15} \\
    \hline 
   6 &Predicting future of polarity of images on social networks \& Images  &Gaussian Naive Bayes &Accuracy 90\% \cite{b17} \\
    \hline 
   7 &To predict the spread of propaganda in social media \& Textual data  &Support Vector Machine &Accuracy 84\% \cite{b24} \\
    \hline 
   8 &A novel sentiment analysis of social networks using supervised learning \& Textual data- Twitter  &Support Vector Machine &Accuracy 88\% \cite{b25} \\
    \hline
   9 &Detecting and identifying hate speech on social media comments Textual data -Twitter, Facebook and Myspace  &Random Forest &Accuracy 72.2\% \cite{b26} \\
    \hline
   10 &Detecting depression Textual data-Twitter  &Random Forest  &Accuracy 74\% \cite{b27} \\
    \hline 
   11 &To analyze sentiments on Indonesia language  &Random Forest  &Accuracy 82.9\% \cite{b28} \\
    \hline 
   12 &To detect cyber-bullying of children - Textual data from Facebook and Twitter  &Random Forest  &Accuracy 83\% \cite{b36} \\
    \hline  
   13 &Ensemble classification of cyber space users tendency in blog writing using Textual data- Facebook, Twitter  &Random Forest  &Accuracy 88\% \cite{b30} \\
    \hline  
   14 &To identify critical indicators of the rise and decline of stock market  &Random Forest  &Accuracy 98.34 88\% \cite{b31} \\
    \hline    
  \end{tabular}}
  \caption{Summary of the reviewed machine learning application cases for sentiment analysis}
\end{table*}

\section{Conclusion and Future Work}
This paper reviewed social network analysis and semantic analysis and associated state-of-the-art techniques. The primary objective of this work is to review various techniques uses in SNA and SA and determine their performance in relation to politics. This work noted that supervised machine learning method such as support vector machine (SVM) and Naïve Bayes are widely used in sentiment analysis. This is due to their simplicity and accuracy \cite{b37}. Classification using Naïve Bayes increases the accuracy score when applied to well-formed textual data \cite{b32}. This work has also observed that applying more than one technique optimizes the performance without compromising the accuracy \cite{b38}. Despite remarkable accuracy score in use of the various techniques, there are grey areas in relation to political influence that future researchers may look at. Such areas are to have an overview of necessary procedures such as data collection, data cleaning and feature selection used in various techniques, expanding the scope of data collected to include images, mentions, hashtags, and to come up with ways of dealing with sarcasm, negated sentences, misspelt words, and slang or unstructured words \cite{b12}.

\vspace{12pt}


\begin{thebibliography}{00}
\bibitem{b1} N. Öztürk and S. Ayvaz, “Sentiment analysis on Twitter: A text mining approach to the Syrian refugee crisis,” Telemat. Informatics, vol. 35, no. 1, pp. 136–147, 2018, doi: 10.1016/j.tele.2017.10.006.
\bibitem{b2} R. Cantini, F. Marozzo, D. Talia, and P. Trunfio, “Analyzing Political Polarization on Social Media by Deleting Bot Spamming,” Big Data Cogn. Comput., vol. 6, no. 1, 2022, doi: 10.3390/bdcc6010003.
\bibitem{b3} A. Khan et al., “Predicting Politician’s Supporters’ Network on Twitter Using Social Network Analysis and Semantic Analysis,” Sci. Program., vol. 2020, 2020, doi: 10.1155/2020/9353120.
\bibitem{b4} M. Nur Habibi and Sunjana, “Analysis of Indonesia Politics Polarization before 2019 President Election Using Sentiment Analysis and Social Network Analysis,” Int. J. Mod. Educ. Comput. Sci., vol. 11, no. 11, pp. 22–30, 2019, doi: 10.5815/ijmecs.2019.11.04.
\bibitem{b5} Z. Drus and H. Khalid, “Sentiment analysis in social media and its application: Systematic literature review,” Procedia Comput. Sci., vol. 161, pp. 707–714, 2019, doi: 10.1016/j.procs.2019.11.174.
\bibitem{b6} A. Hasan, S. Moin, A. Karim, and S. Shamshirband, “Machine Learning-Based Sentiment Analysis for Twitter Accounts,” Math. Comput. Appl., vol. 23, no. 1, p. 11, 2018, doi: 10.3390/mca23010011.
\bibitem{b7} M. Wongkar and A. Angdresey, “Sentiment Analysis Using Naive Bayes Algorithm Of The Data Crawler: Twitter,” in 2019 Fourth International Conference on Informatics and Computing (ICIC), 2019, pp. 1–5, doi: 10.1109/ICIC47613.2019.8985884.
\bibitem{b8} A. Kruseman, “Social Network Analysis on Tweets,” 2018. .
\bibitem{b9} B. T.k., C. S. R. Annavarapu, and A. Bablani, “Machine learning algorithms for social media analysis: A survey,” Comput. Sci. Rev., vol. 40, p. 100395, 2021, doi: 10.1016/j.cosrev.2021.100395.
\bibitem{b10} J. Kim, S. Hwang, and E. Park, “Can we predict the Oscar winner? A machine learning approach with social network services,” Entertain. Comput., vol. 39, no. August 2020, p. 100441, 2021, doi: 10.1016/j.entcom.2021.100441.
\bibitem{b11} M. Ansari, M. B. Aziz, M. O. Siddiqui, H. Mehra, and K. P. Singh, “Analysis of Political Sentiment Orientations on Twitter,” Procedia Comput. Sci., vol. 167, pp. 1821–1828, Jan. 2020, doi: 10.1016/j.procs.2020.03.201.
\bibitem{b12} H. Soong, N. B. A. Jalil, R. K. Ayyasamy, and R. Akbar, “The Essential of Sentiment Analysis and Opinion Mining in Social Media,” 2019 IEEE 9th Symp. Comput. Appl. Ind. Electron., pp. 272–277, 2019.
\bibitem{b13} N. Kewsuwun and S. Kajornkasirat, “A sentiment analysis model of agritech startup on Facebook comments using naive Bayes classifier.,” Int. J. Electr. \& Comput. Eng., vol. 12, no. 3, 2022.
\bibitem{b14} V. A. Fitri, R. Andreswari, and M. A. Hasibuan, “Sentiment analysis of social media Twitter with case of Anti-LGBT campaign in Indonesia using Naïve Bayes, decision tree, and random forest algorithm,” Procedia Comput. Sci., vol. 161, pp. 765–772, 2019, doi: 10.1016/j.procs.2019.11.181.
\bibitem{b15} M. M. Itani, R. N. Zantout, L. Hamandi, and I. Elkabani, “Classifying sentiment in Arabic social networks: Naïve search versus Naïve Bayes,” 2012 2nd Int. Conf. Adv. Comput. Tools Eng. Appl. ACTEA 2012, pp. 192–197, 2012, doi: 10.1109/ICTEA.2012.6462864.
\bibitem{b16} A. Filzah, M. Nasir, K. Amin, and M. Sukri, “Journal of Soft Computing and Data Mining Machine Learning Approach on Cyberstalking Detection in Social Media Using Naive Bayes and Decision Tree,” J. Soft Comput. Data Min., vol. 3, no. 1, pp. 19–27, 2022, [Online]. Available: http://penerbit.uthm.edu.my/ojs/index.php/jscdm.
\bibitem{b17} K. Almgren, J. Lee, and M. Kim, “Predicting the future popularity of images on social networks,” ACM Int. Conf. Proceeding Ser., 2016, doi: 10.1145/2955129.2955154.
\bibitem{b18} M. Kolomeets, A. Chechulin, and I. Kotenko, “Social networks analysis by graph algorithms on the example of the V Kontakte social network,” J. Wirel. Mob. Networks, Ubiquitous Comput. Dependable Appl., vol. 10, no. 2, pp. 55–75, 2019, doi: 10.22667/JOWUA.2019.06.30.055.
\bibitem{b19} S. P. B, “Improved BSP Clustering Algorithm for Social Network Analysis,” Bonfring Int. J. Softw. Eng. Soft Comput., vol. 1, no. 1, pp. 15–20, 2011, doi: 10.9756/bijsesc.1003.
\bibitem{b20} J. R. Sanchez et al., “On the Power of Social Networks to Analyze Threatening Trends,” IEEE Internet Comput., vol. 26, no. 2, pp. 19–26, 2022, doi: 10.1109/MIC.2022.3154712.
\bibitem{b21} C. Udanor, S. Aneke, and B. O. Ogbuokiri, “Determining social media impact on the politics of developing countries using social network analytics,” Program, vol. 50, no. 4, pp. 481–507, Jan. 2016, doi: 10.1108/PROG-02-2016-0011.
\bibitem{b22} A. J. O’Malley and P. V. Marsden, “The analysis of social networks,” Heal. Serv. Outcomes Res. Methodol., vol. 8, no. 4, pp. 222–269, 2008, doi: 10.1007/s10742-008-0041-z.
\bibitem{b23} I. V Kozitsin et al., “Modeling Political Preferences of Russian Users Exemplified by the Social Network Vkontakte,” Math. Model. Comput. Simulations, vol. 12, no. 2, pp. 185–194, 2020, doi: 10.1134/S2070048220020088.
\bibitem{b24} A. M. U. D. Khanday, Q. R. Khan, and S. T. Rabani, “SVMBPI: support vector machine-based propaganda identification,” in Cognitive Informatics and Soft Computing, Springer, 2021, pp. 445–455.
\bibitem{b25} M. Anjaria and R. M. R. Guddeti, “A novel sentiment analysis of social networks using supervised learning,” Soc. Netw. Anal. Min., vol. 4, no. 1, pp. 1–15, 2014, doi: 10.1007/s13278-014-0181-9.
\bibitem{b26} K. Nugroho et al., “Improving Random Forest Method to Detect Hatespeech and Offensive Word,” in 2019 International Conference on Information and Communications Technology (ICOIACT), 2019, pp. 514–518, doi: 10.1109/ICOIACT46704.2019.8938451.
\bibitem{b27} J. Angskun, S. Tipprasert, and T. Angskun, “Big data analytics on social networks for real-time depression detection,” J. Big Data, vol. 9, no. 1, p. 69, 2022, doi: 10.1186/s40537-022-00622-2.
\bibitem{b28} M. A. Fauzi, “Random forest approach fo sentiment analysis in Indonesian language,” Indones. J. Electr. Eng. Comput. Sci., vol. 12, no. 1, pp. 46–50, 2018, doi: 10.11591/ijeecs.v12.i1.pp46-50.
\bibitem{b29} Patel, “Evaluating the Effectiveness of an Ensemble Random Forest Machine Learning Algorithm in Detecting Cyberbullying in the 4chan Politically Incorrect Board Social,” pp. 9–25, 2019.
\bibitem{b30} N. A. Samsudin, A. Mustapha, and M. H. Abd Wahab, “Ensemble classification of cyber space users tendency in blog writing using random forest,” in 2016 12th International Conference on Innovations in Information Technology (IIT), 2016, pp. 1–4, doi: 10.1109/INNOVATIONS.2016.7880046.
\bibitem{b31} M. N. Elagamy, C. Stanier, and B. Sharp, “Stock market random forest-text mining system mining critical indicators of stock market movements,” 2nd Int. Conf. Nat. Lang. Speech Process. ICNLSP 2018, pp. 1–8, 2018, doi: 10.1109/ICNLSP.2018.8374370.
\bibitem{b32} A. U. Hassan, J. Hussain, M. Hussain, M. Sadiq, and S. Lee, “Sentiment Analysis of Social Networking Sites (SNS) Data using Machine Learning Approach for the Measurement of Depression,” 2017, doi: 10.1109/ICTC.2017.8190959.
\bibitem{b33} S. Mahtab, N. Islam, and M. M. Rahaman, “Sentiment Analysis on Bangladesh Cricket with Support Vector Machine,” 2018, pp. 1–4, doi: 10.1109/ICBSLP.2018.8554585.
\bibitem{b34} K. Chekima and R. Alfred, “Sentiment Analysis of Malay Social Media Text,” 2018, pp. 205–219.
\bibitem{b35} C. Dhaoui, C. M. Webster, and L. P. Tan, “Social media sentiment analysis: lexicon versus machine learning,” J. Consum. Mark., vol. 34, no. 6, pp. 480–488, Jan. 2017, doi: 10.1108/JCM-03-2017-2141.
\bibitem{b36} C. C. Henry, “Evaluating the Effectiveness of an Ensemble Random Forest Machine Learning Algorithm in Detecting Cyberbullying in the 4chan Politically Incorrect Board Social,” vol. 15, no. 2, pp. 1–23, 2019.
\bibitem{b37} M. Wankhade, A. Rao, and C. Kulkarni, “A survey on sentiment analysis methods, applications, and challenges,” Artif. Intell. Rev., pp. 1–50, 2022, doi: 10.1007/s10462-022-10144-1.
\bibitem{b38} P. K. Gikunda and N. Jouandeau, State-of-the-Art Convolutional Neural Networks for Smart Farms: A Review, vol. 997. Springer International Publishing, 2019.
\end{thebibliography}
\end{document}